\documentclass[conference]{IEEEtran}
\IEEEoverridecommandlockouts
\usepackage{cite}
\usepackage{amsmath,amssymb,amsfonts}
\usepackage{algorithmic}
\usepackage{graphicx}
\usepackage{textcomp}
\usepackage{xcolor}
\usepackage{multirow}
\usepackage{subfigure}
\def\BibTeX{{\rm B\kern-.05em{\sc i\kern-.025em b}\kern-.08em
    T\kern-.1667em\lower.7ex\hbox{E}\kern-.125emX}}
    \usepackage[left=0.68in,right=0.68in,top=0.75in,bottom=1.1in]{geometry}
\begin{document}

\title{D-Score: A White-Box Diagnosis Score for CNNs Based on Mutation Operators

}

\author{\IEEEauthorblockN{1\textsuperscript{st} Xin Zhang}
\IEEEauthorblockA{
\textit{University of South Carolina}\\
Columbia, United States \\
xz8@email.sc.edu}
\and

\IEEEauthorblockN{2\textsuperscript{nd} Yuqi Song}
\IEEEauthorblockA{
\textit{University of South Carolina}\\
Columbia, United States \\
yuqis@email.sc.edu}
\and

\IEEEauthorblockN{3\textsuperscript{rd} Xiaofeng Wang}
\IEEEauthorblockA{
\textit{University of South Carolina}\\
Columbia, United States \\
wangxi@cec.sc.edu}
\and

\IEEEauthorblockN{4\textsuperscript{th} Fei Zuo}
\IEEEauthorblockA{
\textit{University of Central Oklahoma}\\
Edmond, United States \\
fzuo@uco.edu}

}
\maketitle



\begin{abstract}
Convolutional neural networks (CNNs) have been widely applied in many safety-critical domains, such as autonomous driving and medical diagnosis.  However, concerns have been raised with respect to the trustworthiness of these models: The standard testing method evaluates the performance of a model on a test set, while low-quality and insufficient test sets can lead to unreliable evaluation results, which can have unforeseeable consequences.
Therefore, how to comprehensively evaluate CNNs and, based on the evaluation results, how to enhance their trustworthiness are the key problems to be urgently addressed.
Prior work has used mutation tests to evaluate the test sets of CNNs.  However, the evaluation scores are black boxes and not explicit enough for what is being tested.
In this paper, we propose a white-box diagnostic approach that uses mutation operators and image transformation to calculate the feature and attention distribution of the model and further present a diagnosis score, namely D-Score, to reflect the model's robustness and fitness to a dataset. We also propose a D-Score based data augmentation method to enhance the CNN's performance to translations and rescalings. Comprehensive experiments on two widely used datasets and three commonly adopted CNNs demonstrate the effectiveness of our approach.
\end{abstract}


\section{Introduction}
\label{intro}
In recent years convolutional neural networks (CNNs) have been increasingly used in safety-critical applications, including medical diagnostics~\cite{litjens2017survey,ozturk2020automated}, autonomous vehicles~\cite{grigorescu2020survey,gupta2021deep}, and military fields~\cite{denos2017deep,kafedziski2018detection}. Despite their impressive success, CNNs still face challenges related to robustness and accuracy. For instance, crashes caused by autonomous cars from Tesla and Google have led to significant losses~\cite{feng2020deepgini}. Therefore, as with traditional software, testing is essential for CNN-based systems, which can effectively identify issues and improve the system's trustworthiness~\cite{zhang2020machine}.


The most common method for evaluating a model is to assess its performance on selected evaluation metrics using a test set~\cite{ma2018deepmutation,zhang2020machine,sun2018testing}. However, this method heavily relies on the quality of the test set. In other words, if most instances in the test set have features similar to those in the training set, the model's testing results will likely be good. Conversely, due to reasons such as biased training data, overfitting, and underfitting, the trained model may show unexpected or incorrect behaviors on a test set that contains many corner cases~\cite{pei2017deepxplore}. Thus, when it comes to safety-critical areas, testing a trained model on an unevaluated test set and making decisions based on the test results can lead to catastrophic consequences.

To evaluate CNNs, researchers have proposed several approaches, which can be divided into two categories. The first category involves introducing the traditional software engineering testing method, mutation testing, to CNNs~\cite{hu2019deepmutation++, humbatova2021deepcrime, shen2018munn}. This approach applies carefully designed mutation operators~\cite{ma2018deepmutation} to the CNN model to generate multiple variants. The higher the number of differences between the predictions of the variant models and the original model, the higher the quality of the test set. However, the score itself remains a black box, and the reasons behind the low quality of the test set are still unknown. 
Additionally, effective methods for selecting and combining mutation operators to detect test set quality remain unexplored~\cite{panichella2021we}.
The second category of approaches is based on neuron coverage~\cite{pei2017deepxplore, feng2020deepgini, yu2019test4deep}. These methods use gradient ascent to solve a joint optimization problem that maximizes both neuron coverage and the number of potentially erroneous behaviors, and eventually generate a set of test inputs~\cite{pei2017deepxplore}. However, as noted in~\cite{harel2020neuron}, higher neuron coverage can lead to fewer defects detected, less natural inputs, and more biased prediction preferences. Therefore, developing effective methods for providing white-box scores for CNNs and proposing methods for enhancing these scores is critical for improving robustness and accuracy of~CNNs.

This paper investigates the issue of how to diagnose CNNs using a white-box approach. To generate several variants, we employ the mutation operator of deleting neurons~\cite{ma2018deepmutation}. Unlike previous work, where random neuron selection was used for deletion~\cite{shen2018munn,humbatova2021deepcrime}, we divide the neurons of each convolutional layer into several regions and delete the neurons of each region, enabling us to study the overall feature distribution of the test set due to the spatial character of CNNs. We then analyze the attention of the CNN model towards different regions by applying well-designed image transformations such as padding for given directions.  Based on the overall feature distribution and attention distribution, we introduce the concept of D-Score for CNNs that reflects their robustness and fitness, where ``robustness'' of a CNN is the ability of the CNN in recognizing objects at any location of an image (e.g., translation invariance) and ``fitness'' of a CNN on a dataset means how well the attention of the CNN meets the feature distribution of the dataset.


Based on the D-Score, we propose a scoring-guided data augmentation strategy to enhance CNN's robustness and fitness.  It is known that CNNs are generally not robust enough to image transformation (e.g., translations and/or rescalings of the input image may drastically change the prediction of a CNN~\cite{small,zhang2019making}).  
Although data augmentation is consistently considered as an effective strategy to address this issue~\cite{shorten2019survey}, our experiments show that randomly and blindly selecting data augmentation techniques with little knowledge on the dataset will significantly limit such effectiveness.
Instead, our data augmentation strategy fully utilizes D-Score, resulting in more targeted selection and design of augmentation techniques as well as their execution probabilities. This approach can effectively adjust the original feature distribution of the dataset, making the trained model more robust and reducing blind spots.

%


The main contributions are summarized as follows:
\begin{itemize}
    \item By analyzing the impact of mutation operators on the accuracy of CNNs through deleting neurons in different regions and applying image transformations, we develop an approach to calculate the overall feature distribution of a dataset, as well as the attention distribution of models. This allows us to white-box diagnose CNNs and introduce a new concept of D-Score for CNN diagnosis, providing valuable insights into their performance.
    \item In order to showcase the efficiency of our D-Score and enhance the robustness and fitness of CNNs, we introduce a score-guided data augmentation approach that tackles the problem of CNNs' sensitiveness to shifts and rescalings.
    \item Our scoring and data augmentation method has been rigorously tested on two widely used datasets and three commonly adopted CNNs for these datasets, with comprehensive experiments confirming its effectiveness.
\end{itemize}

The rest of the paper is organized as follows.  Section~\ref{rw} discusses the related work.  Section~\ref{f} introduces some preliminaries in CNNs.  Our proposed method is presented in Section~\ref{method}.  Section~\ref{exper} shows the experimental results.  Finally, conclusions are drawn in Section~\ref{con}.







\section{Related Work}
\label{rw}
In this section, we will briefly introduce mutation testing (MT) on classical software and then discuss the state-of-the-art testing techniques to evaluate CNNs. 

\subsection{Traditional Mutation Testing}
\label{sub:classical}
MT was first proposed in~\cite{offutt2001mutation} and became a popular method to assess the quality of test suites~\cite{ma2018deepmutation, ma2006mujava,papadakis2019mutation}. 
The basic idea of MT is to inject artificial faults into the production code by applying mutation operators~\cite{agrawal1989design} such that a set of faulty program mutants can be generated. 
For instance, changing the $>$ operator in the original program $P:if~(a>b) ~return ~True$ into $<$ and generating a mutant program $P_m:if~(a<b) ~return ~True$. 
For a test suite $t$, it kills the mutant $P_m$ if it receives different outputs when executed against $P$ and $P_m$ individually. 
The more mutants $t$ kills, the higher quality the software has~\cite{jia2010analysis}.

\subsection{Mutation Testing for Deep Learning}
\label{sub:mt}
The growing popularity of deep learning has raised concerns about the robustness and reliability of deep neural networks (DNNs), leading to a rise in research interest in mutation testing for deep learning~\cite{hu2019deepmutation++,humbatova2021deepcrime,ma2018deepmutation,panichella2021we}. However, unlike traditional software, where the decision logic is coded by developers, the behavior of deep learning systems is determined by the structures of DNNs and the parameters in the network~\cite{wang2019adversarial} which is hard to foresee. Moreover, due to the randomness of the training process, it is common to observe different decisions when a DNN is retrained on the same dataset, even without any mutation operations. This makes it difficult to apply the mutation-killing metric of traditional software to learning systems directly~\cite{jahangirova2020empirical}.


In order to make the application of mutation testing in deep learning systems feasible, several different approaches have been proposed. MuNN~\cite{shen2018munn} proposes five mutation operators, including replacements or deletions of neurons, activation functions, and parameters on trained CNNs. DeepMutation~\cite{ma2018deepmutation} focuses on mutation at both source-level and model-level, generating two types of operators to mutate the training data, model structures during and after training, and model parameters. DeepMutation++\cite{hu2019deepmutation++} expands the prior work from CNN to RNN. DeepCrime\cite{humbatova2021deepcrime} defines mutation operators based on studies of real faults in learning systems. Furthermore, the mutation-killing metric in deep learning systems is discussed in detail in~\cite{jahangirova2020empirical}.
%
%
As pointed out in~\cite{panichella2021we}, however, the use of these methods for evaluating a test set can only reveal the number of variant models that can be discovered by this test set. These scores do not provide clear insight into the trustworthiness of the model. Therefore, these scores remain black boxes and are not sufficiently explicit in terms of what exactly is being tested.


\subsection{Neuron Coverage Based Test}
\label{sub:nc}

Neuron coverage (NC) is a test-based evaluation approach that provides a different perspective for evaluating the test adequacy of deep learning systems. NC is defined as the ratio of the number of activated neurons, whose outputs are greater than a specified threshold, to the total number of neurons. The NC value is different for each test input. This approach has been widely used to improve the performance of deep learning systems. For instance, DeepXplore~\cite{pei2017deepxplore} uses gradient ascent to maximize both neuron coverage and the number of potentially erroneous behaviors based on multiple DNNs. DeepGini~\cite{feng2020deepgini} introduces several NC criteria and proposes a test prioritization method based on a statistical perspective of DNNs. Test4Deep~\cite{yu2019test4deep} focuses on a single DNN and induces inconsistencies between the predicted labels of original inputs and those of generated test inputs. However, extensive evaluation has shown that increasing NC can make it harder to generate an effective test suite. Higher neuron coverage leads to fewer detected defects, less natural inputs, and more biased prediction preferences~\cite{harel2020neuron,gannamaneni2022good}.

%


\section{Preliminaries}
\label{f}
A CNN usually consists of three types of layers: convolutional layers, pooling layers, and fully-connected layers. 
When these layers are stacked, a CNN architecture has been formed~\cite{albawi2017understanding}. Here we briefly discuss the spatial characteristics and translational variance of CNNs, which are crucial to the design of our approach.

\subsection{Spatial Characteristics }

\begin{figure}[ht]
  \centering
  \vspace{-3mm}
   \includegraphics[width=0.45\textwidth]{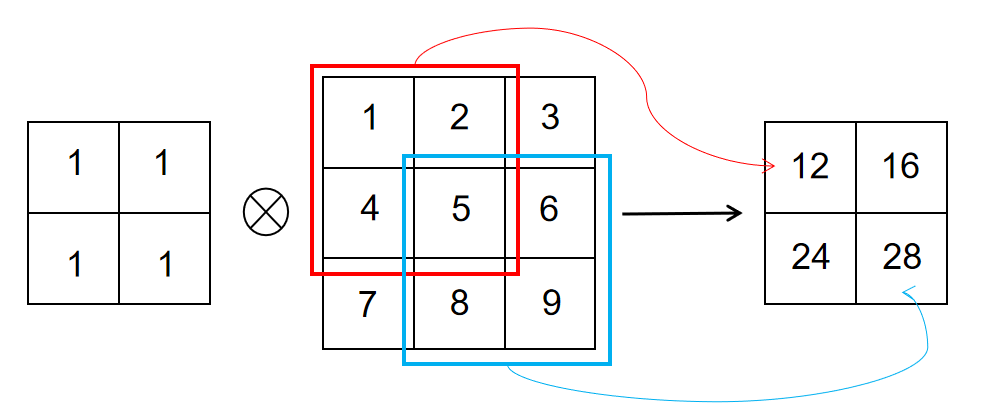}
     \vspace{-3mm}
    \caption{Illustration of the convolution operation.}
      \vspace{-3mm}
	\label{fig:conv}
\end{figure}

The spatial characteristics of CNNs mainly originate from the way convolution is calculated. CNNs rely on convolution kernels, which are typically small in spatial dimensionality but spread throughout the entirety of the input following a left-to-right, top-to-bottom order, to convolve the input matrices~\cite{o2015introduction}. When the data hits a convolutional layer, the layer convolves each filter across the spatial dimensionality of the input to produce a 2D activation map. As a result, the relative positions of the features in the input are maintained after being convolved~\cite{wu2017introduction}. For example, as shown in Figure~\ref{fig:conv}, the features of the region with blue color still appear in the right-bottom region after being convolved.

\subsection{Translation Variance }


CNNs are often assumed to be invariant to small image transformations in theory~\cite{fukushima1980neocognitron,fukushima1982neocognitron}. However, recent studies have shown that this is not always the case. Small translations or rescalings of the input image can significantly alter the network's prediction~\cite{sp,small}. This issue can arise due to the concentration of features in the dataset~\cite{small} or the insufficient network architecture~\cite{zhang2019making}.
If a CNN is robust, it should be able to recognize the object wherever it appears in the image.  In this sense, the lack of translation invariance will be considered as a manifestation of the model's poor robustness, which indicates that the current model may not accurately recognize certain corner cases.

\section{Methodology}
\label{method}
This section introduces the concept of D-Score for a CNN, which is generated based on mutation operations and image transformations. We then show how to use D-Score to enhance the robustness and fitness of the model. 


\subsection{D-Score}
\label{sub:score}

\begin{figure*}[ht]
  \centering
   \includegraphics[width=0.96\textwidth]{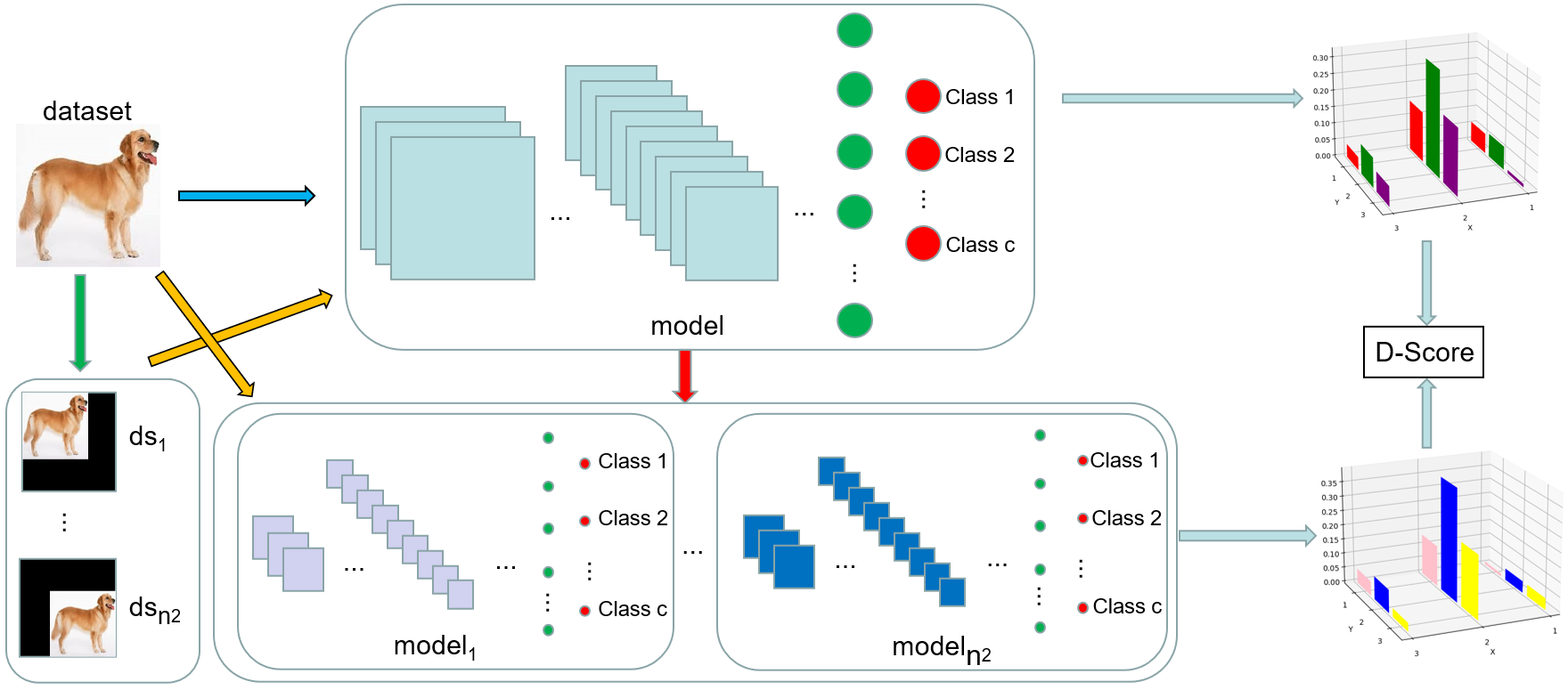}
     \vspace{-3mm}
    \caption{The pipeline of D-Score calculation. Given a well-trained CNN (labeled as ${\bf model}$), we generate $n^2$ model variants and $n^2$ new test sets using mutation operators and image transformations, respectively, and obtain ${\bf model}_1,\ldots, {\bf model}_{n^2}$ and $ds_1,\ldots, ds_{n^2}$. Next, we obtain the distribution of features and attention through the accuracy of $n^2$ model variants on the original test set and the accuracy of the original model on $n^2$ new test sets, respectively, based on which we calculate D-Score of ${\bf model}$.}
      \vspace{-3mm}
	\label{fig:pipeline}
\end{figure*}

Models and data are two critical components of deep learning. 
On the one hand, the data effectively guides the model to learn. On the other hand, the model can accurately predict the data.  To ensure the impartiality and representativeness of the proposed D-Score, let us start our scoring method with the interplay between the data and the model.  Specifically, our method takes the following two aspects into consideration:
\begin{itemize}
\item Keep the dataset fixed and evaluate it through changing the model.
By removing different regions of neurons through mutation operations, we analyze the impact of specific regions of neurons on the predicted results. This essentially leads to an analysis of the feature distribution of the dataset, with the consideration of the spatial characteristics of CNNs~\cite{albawi2017understanding}.
\item Keep the model fixed and evaluate it through changing the dataset.
To study how the variations in the locations of the target in images will affect model predictions, we change the dataset by carefully designing the image transformations, which involve padding the original images in a specific direction and resizing them. By doing so, we can obtain the attention distribution of the CNN.
\end{itemize}
Performing a comprehensive analysis of these two distributions, we can calculate D-Score for the evaluated CNN.
The overall pipeline of our scoring approach is shown in Fig.~\ref{fig:pipeline}.

\begin{figure}[ht]
  \centering
   \includegraphics[width=0.49\textwidth]{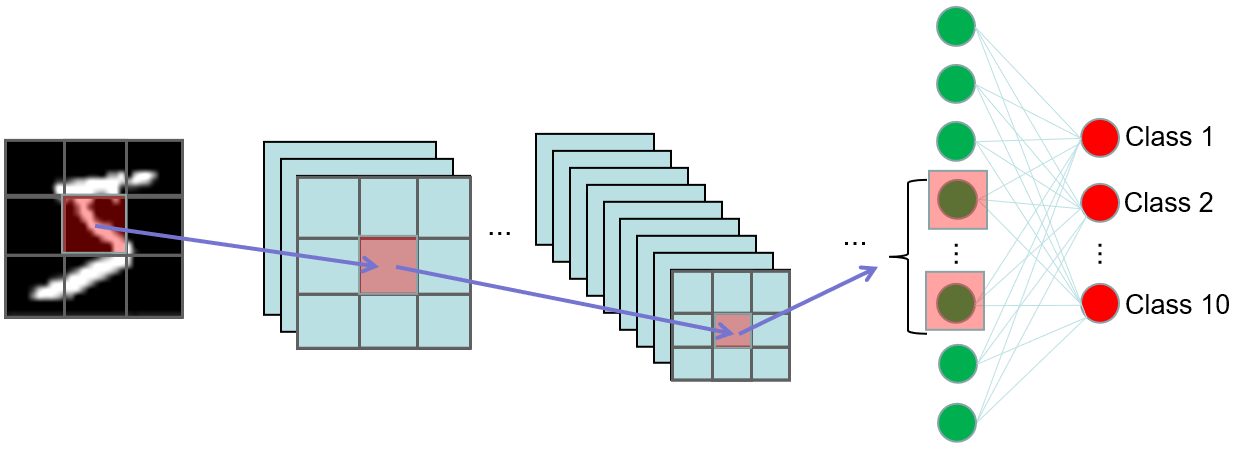}
     \vspace{-5mm}
    \caption{Using deleting mutation operator to generate variants. In this case, we divide each convolutional layer into $3\times 3$ regions, and then apply deleting operator on the region with index 5. 
    The red masks in the convolutional layers indicate that the corresponding neurons in the $5$th region will be deleted, which will result in the loss of input information at the fully connected layer.  By ``deleting a region of neurons'', it means that this deletion operation is applied to the same region of every convolutional layer, which will clearly result in the loss of features from certain regions of the input images.}
      \vspace{-4mm}
	\label{fig:deleteN}
\end{figure}

\smallskip
\noindent
\textbf{Mutation Operators.}
There exist different mutation operators, such as deleting neurons, deleting layers, adding layers, and changing activation functions, to name a few~\cite{ma2018deepmutation,panichella2021we}.
Since our goal is to first analyze the feature distribution of the dataset based on the spatial characteristics of the CNN, we only adopt the mutation operators for deleting neurons~\cite{shen2018munn} to generate variants.  It is worth mentioning that in the existing approaches, the neurons to be deleted are usually chosen randomly since they focus more on identifying variants. However, random deletion cannot meet our requirements for the purpose of analyzing the feature distribution of the dataset. Thus, we propose a region-based method to delete the neurons: First, divide the neurons of each convolutional layer into $n\times n$ equal-sized rectangular regions, where $n$ is a hyperparameter, and index them in order from the upper-left corner to the lower-right corner before performing the deletion operation; Then, delete the corresponding regions of neurons with the same index in all convolutional layers from the well-trained CNN model to form $n^2$ variants, as shown in Fig.~\ref{fig:deleteN}. This approach can prevent introducing the surrounding features into the deleted region (e.g., due to operations such as pooling) for a more accurate feature distribution of the~dataset.

\smallskip
\noindent
\textbf{Feature Distribution.}
We take the original well-trained model as the baseline and calculate the accuracy difference between the variants and the baseline. Clearly, the larger the difference is, the worse performance the variant has, and the more important the related region of neurons is. Due to the spatial characteristics of the CNN, deleting the $i$th region of neurons implies loss of the information of the input image at a corresponding region when predicting.  In this sense, the difference in accuracy of the $i$th variant model from the baseline can be used to represent the feature quantity of the dataset in that specific region.  So we normalize the accuracy differences of the $n^2$ variants from the baseline to obtain the feature distribution of the dataset over regions:
\begin{equation} 
\widetilde{f}_i = \frac{\max(f_b-f_i,0)}{\sum_{i=1}^{n^2} \max(f_b-f_i,0)}, \label{norm}
\end{equation}
where $\widetilde{f}_i$ and $f_i$ stands for the value of the $i$-th region after and before normalization, respectively, and $f_b$ represents the accuracy of baseline model.  



\begin{figure}[ht]
  \centering   \vspace{-3mm}
   \includegraphics[width=0.4\textwidth]{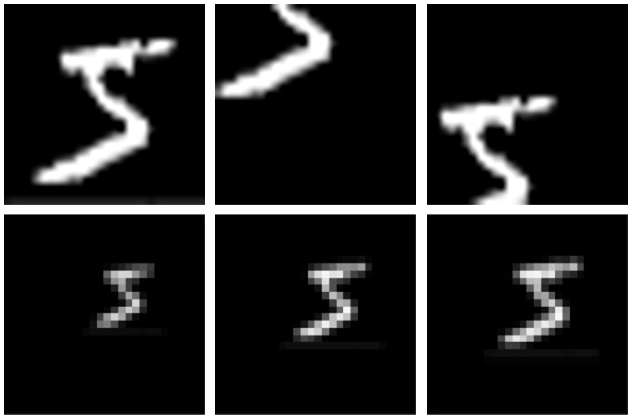}
     \vspace{-2mm}
    \caption{Image transformation. The first row includes the original image, followed by two images after traditional translation operations, which demonstrate a significant loss of information.  The second row shows the images after applying our image transformation method with $t=2,3,4$, respectively.}
      \vspace{-2mm}
	\label{fig:imageT}
\end{figure}

\smallskip
\noindent
\textbf{Image Transformation.}
To evaluate a model, we propose an image ``translation'' method to modify the dataset, which is different from the traditional translation that discards features outside the field of view (e.g., two sub-figures at the right side of the top row in Fig.~\ref{fig:imageT}).  In our approach, we first divide the original image into $n\times n$ equal-sized regions and index them in order from the upper-left corner to the lower-right corner. Then we push the original image towards the target region by padding 0 around the original image. To be more specific, given the original image size $k \times l$ and a hyperparameter $t$, we fill $\frac{l}{t}$, $\frac{2l}{t}$, $\frac{2k}{t}$ and $\frac{k}{t}$ blank pixels in the top, bottom, left, and right directions of the original image, respectively, and then resize the newly generated image back to its original size so that it fits the dimensions of the trained model. The purpose of introducing $t$ is to prevent the effective area of the image from being too small after transformation; otherwise, it may lead to low prediction accuracy. For example, the bottom row in Fig.~\ref{fig:imageT} shows the translated images with different values of $t$, given $n=4$ (16 regions in total) and the target region $7$.  With image transformation, we obtain $n^2$ new test sets.

\smallskip
\noindent
\textbf{Attention Distribution.}
We use the original trained model to predict the results on the newly generated $n^2$ test sets and then use the model's accuracy on the $i$th test set, as the indicator of the model's attention to the $i$th region of an image. Intuitively, the better the model performs on the $i$th test set, the more attention it pays to the $i$th region.
To make it consistent with the feature distribution of datasets, we normalize the attention to obtain the attention distribution:
\begin{equation}
\widetilde{a}_i = \frac{a_i}{\sum_{i=1}^{n^2}a_i},  \label{norma} 
\end{equation}
where $a_i$ and $\widetilde{a}_i$ stands for the accuracy before and after normalization, respectively. 


\smallskip
\noindent
\textbf{D-Score Calculation.}
The similarity between the feature distribution and the attention distribution is crucial for successful learning. When these two distributions are close, namely that the model can focus exactly on the feature-dense regions of the dataset, the model's accuracy will be excellent on this dataset. Otherwise, it means that the model cannot capture the features of the dataset well.  Therefore, $\frac{1}{n^2}\sqrt{\sum_{i=1}^{n^2}(\widetilde{f}_i-\widetilde{a}_i)^2}$ can, to some extent, reflect the accuracy degradation. Keeping this in mind, we define the fitness of a CNN on a dataset as
\begin{equation}
v_{\rm fitness} = \hat{a}-\frac{1}{n^2}\sqrt{\sum_{i=1}^{n^2}(\widetilde{f}_i-\widetilde{a}_i)^2}, \label{eq:acc}
\end{equation}
where $\hat{a}$ stands for the accuracy of the original trained model on the original test set.

To define the robustness index, let $a_{\rm avg} = \frac{1}{n^2}$, which represents the average attention distribution of the model over $n^2$ regions, i.e., the model pays equal attention to each region of an image. So the difference between $\widetilde a_i$ and $a_{\rm avg}$ represents the unbalanced attention of the model on the $i$th region. Similarly, let $f_{\rm avg} = \frac{1}{n^2}$ be the average feature distribution of the dataset. We can define the robustness index~as
\begin{align}
    v_{\rm robust} = &~\frac{1}{n^2} \left(\sqrt{\sum_{i=1}^{n^2}(\widetilde{f}_i-f_{\rm avg})^2} + \sqrt{\sum_{i=1}^{n^2}(\widetilde{a}_i-a_{\rm avg})^2}\right. \nonumber \\& + \left.\sqrt{\sum_{i=1}^{n^2}(a_i-\hat{a})^2}\right).  
    \label{eq:vr}
\end{align}
Obviously, if the first two terms in the right side of equation~\eqref{eq:vr} are large, it indicates that the model's attention is concentrated on specific regions, reflecting the model's sensitivity to translation and scaling.  The third term, $\sqrt{\sum_{i=1}^{n^2}(a_i-\hat{a})^2})$, measures the difference of model accuracy when images appear in different regions from the accuracy of the original model on the original dataset. A large value of this term means that the model has a poor ability in handling corner cases.

Notice that $v_{\rm robust}$ is bounded. Since $\widetilde{f}_i \in [0,1]$, we have 
\begin{align*}
\frac{1}{n^2}\sqrt{\sum_{i=1}^{n^2}(\widetilde{f}_i-f_{\rm avg})^2} & \le   \frac{1}{n^2}\sqrt{(1-\frac{1}{n^2})^2 + (n^2-1)(\frac{1}{n^2})^2} \\
& = \frac{\sqrt{n^2-1}}{n^3}.
\end{align*}
The maximum is achieved when $\widetilde f_i =1$ for a specific $i$ and the other $\widetilde f_i = 0$.  Similarly, we have
\[
\frac{1}{n^2}\sqrt{\sum_{i=1}^{n^2}(\widetilde{a}_i-a_{\rm avg})^2} \le \frac{\sqrt{n^2-1}}{n^3}.
\]
Let $c$ denote the number of classes in the dataset. The worst accuracy of the original model and its variants is $\frac{1}{c}$ in a probabilistic sense, namely picking classes randomly.  So the smallest value for $a_i$ and $\hat a$ is $\frac{1}{c}$. Then 
\[
\frac{1}{n^2} \sqrt{\sum_{i=1}^{n^2}(a_i-\hat{a})^2}
\le \frac{1}{n^2} \sqrt{n^2(1-\frac{1}{c})^2} = \frac{1}{n}\frac{c-1}{c}.  
\]
Therefore, with the inequalities above, we have $v_{\rm robust} \le g(n)$ where 
\begin{align} \label{eq:fn}
g(n) = \frac{2\sqrt{n^2-1}}{n^3} + \frac{1}{n}\frac{c-1}{c}.
\end{align}

With $v_{\rm fitness}$ and $v_{\rm robust}$, we can define D-Score as
\begin{align}
     \text{D-Score} = v_{\rm fitness}-v_{\rm robust} .\label{score}
\end{align}
Notice that a small $v_{\rm robust}$ and a large $v_{\rm fitness}$ are expected, which will result in a large D-Score, meaning that the model achieves high robustness and fitness.  
The efficiency of D-Score will be demonstrated in the next section through experiments.

\subsection{Score-Guided Augmentation}
To demonstrate the effectiveness of D-Score in reflecting the robustness of CNNs, we propose a score-guided method to address the problem of CNNs' insensitivity to translations and rescalings (i.e., when an image is translated or rescaled, the performance of CNN decreases)~\cite{small,sp,zhang2019making}.
To address this issue, a common solution is to locate and adjust objects to be detected in input images before performing predictions, such as adding spatial transformations~\cite{sp} before CNNs.  Essentially, this solution aims to change the feature distribution of datasets, but requires training or other geometric methods to gain prior knowledge. Here we remove this requirement of prior knowledge using D-Score.

Our method primarily utilizes the image transformation approach described in Section~\ref{sub:score}, which involves adding empty values around the image and resizing it.
It is critical in this method to determine the execution probability $p$, image size after padding $d$, and the number of empty values to be padded in the four directions (left, right, up, down), denoted as $d_l$, $d_r$, $d_u$, and $d_d$.
%
Our design is inspired by the idea that a model with lower robustness needs to apply this data augmentation technique with higher probability and a broader range to enhance its robustness. Since $v_{\rm robust}$ is an effective measure of the model's robustness, we utilize it to directly determine both $p$ and $d$.
In our method, the probability of execution and the size after padding are determined by the model's robustness score, namely that
\begin{equation} \label{eq:p}
p=\frac{v_{\rm robust}}{g(n)},~~~d = (1+p) \times \hat{d},
\end{equation}
where $g(n)$ is defined in~\eqref{eq:fn}, $\hat{d}$ stands for the original size.  Notice that $p \in [0,1]$ since $g(n)$ is an upper bound on $v_{\rm robust}$.
%
%

It is worth mentioning that $p=1$ means $v_{\rm robust} = g(n)$, which implies the worst robustness. It is corresponding to the case where all features of the dataset are concentrated in one region ($\widetilde f_{i^*} = 1$ for a specific $i^*$ and $\widetilde f_{i} = 0$ for other $i$s). Similarly, the attention of the model focuses on one specific region ($\widetilde a_{j^*} = 1$ for a specific $j^*$ and $\widetilde a_{j} = 0$ for other $j$s). Therefore, data augmentation is imperative in this case.

We calculate $d_l$ and $d_r$ as follows:
\begin{equation}
\begin{aligned}
d_l &= {\rm random}(0, p \times \hat{d}),\\
d_r &=  p \times \hat{d} - d_l, \label{cald},
    \end{aligned}
\end{equation}
where ${\rm random}$ means the uniform distribution over $[0, p \times \hat{d}]$.
Similarly, we can obtain $d_u$ and $d_d$.

\section{Experiments}
\label{exper}

\subsection{Models and Datasets}
\label{sub:ds}
The experiments are conducted based upon two datasets: MNIST~\cite{deng2012mnist} and CIFAR-10~\cite{krizhevsky2017imagenet}. 
MNIST is a dataset used for handwritten digit image recognition, which includes 60,000 training samples and 10,000 test samples in totally 10 classes (digits from 0 to 9).
The CIFAR-10 dataset is a collection of images used for general-purpose image classification, including 50,000 training samples and 10,000 test samples in 10 different classes (such as airplanes, cars, birds, and cats).

For MNIST, we consider two widely used CNN models proposed in~\cite{lecun1998gradient, xiao2018generating}.   
For CIFAR-10, we use the CNN model in~\cite{carlini2017towards}.
The structures of these CNN models are summarized in Table~\ref{table:model}.
As suggested in~\cite{ma2018deepmutation}, 
We follow the instructions described in~\cite{lecun1998gradient,xiao2018generating,carlini2017towards} to train these three models. 
After training, the MNIST model A (MMA) achieves an accuracy of 98.56\% and an average loss of 0.0413 on the test set, while the MNIST model B (MMB) achieves an accuracy of 99.08\% and an average loss of 0.0149\%, representing the state-of-the-art performance.
For the CIFAR-10 model (CM), its accuracy on the training set can reach 98.01\%, while only 79.66\% on the test set with an average loss of 0.7049.
The performance of these three models is nearly identical to~\cite{ma2018deepmutation}.

\begin{table}[h]
\caption{The structures of our selected CNN models, which are widely adopted for MNIST and CIFAR-10 in the prior work. We use these three models as baselines and apply mutation operators of deleting to generate variants.}
\begin{center}
\begin{tabular}{|l|l|l|}
\hline
\multicolumn{1}{|c|}{MNIST model A~\cite{lecun1998gradient}} & \multicolumn{1}{c|}{MNIST model B~\cite{xiao2018generating}} & \multicolumn{1}{c|}{CIFAR-10 model~\cite{carlini2017towards}} \\ \hline
Conv(6,5,5)+ReLU()                  & Conv(32,3,3)+ReLU()                & Conv(64,3,3)+ReLU()                 \\
MaxPooling(2,2)                     & Conv(32,3,3)+ReLU()                & Conv(64,3,3)+ReLU()                 \\
Conv(16,5,5)+ReLU()                 & MaxPooling(2,2)                    & MaxPooling(2,2)                     \\
MaxPooling(2,2)                     & Conv(64,3,3)+ReLU()                & Conv(128,3,3)+ReLU()                \\
Flatten()                           & Conv(64,3,3)+ReLU()                & Conv(128,3,3)+ReLU()                \\
FC(120)+ReLU()                      & MaxPooling(2,2)                    & MaxPooling(2,2)                     \\
FC(84)+ReLU()                       & Flatten()                          & Flatten()                           \\
FC(10)+Softmax()                    & FC(200)+ReLU()                     & FC(256)+ReLU()                      \\
                                    & FC(10)+Softmax()                   & FC(256)+ReLU()                      \\
                                    &                                    & FC(10)+Softmax()                    \\ \hline
\end{tabular}
\label{table:model}
\end{center}
\end{table}

\subsection{Implementation Details}
\label{sub:detail}
\smallskip
\noindent
\textbf{Deleting Mutation Operator.}
Performing neuron deletion operation does not mean removing the target neuron, but rather blocking the neuron's effect on the following layers so that it has no effect on the subsequent neurons. So we simply set the convolution results for the target neurons to 0 during the forward propagation of the model's predictions on the test sets, which means that these ``deleted'' neurons will not have impacts on final predictions. 

\smallskip
\noindent
\textbf{Image Translation.}
To avoid losing features that move out of the frame when shifting the image, we apply the pad transformer to add padding to all sides of the image with specified values. Afterward, we use the resize transformer to resize the padded image to its original size, ensuring that the trained CNN can make predictions directly without the need of adjusting the input layer's dimensions.

\smallskip
\noindent
\textbf{Methods for Comparison.}
Though there exist methods that can score a test set~\cite{shen2018munn,ma2018deepmutation,hu2019deepmutation++}, they cannot be applied to evaluate CNNs.  To the best of our knowledge, there are few methods that can be directly compared to D-Score for evaluating CNNs.
To demonstrate the effectiveness of our scoring-guided augmentation method, we compare it with several other commonly used data augmentation methods listed in Table~\ref{table:othermethods}.
It is worth pointing out that some augmentation operations can result in a change of meaning for the MNIST data due to its specificity. 
For instance, vertical flipping can cause a digit 6 to become a digit 9. 
Hence, we only apply Random Padding + Resize (RPR) to the MNIST dataset.
\begin{table}[]
\caption{The data augmentation methods adopted for comparisons.}
\begin{tabular}{|c|l|l|}
\hline
\multicolumn{1}{|l|}{Dataset} & Method                            & Parameters               \\ \hline
\multirow{4}{*}{CIFAR-10}     & Random Horizontal Flip (RHF)           & prob=0.5            \\ \cline{2-3} 
                              & Random Vertical Flip (RVF)             & prob=0.5            \\ \cline{2-3} 
                              & Random Rotation (RR)                  & degree in (0,180)   \\ \cline{2-3} 
                              & Random Hor + Ver Flip (RHV) & prob=0.5            \\ \cline{2-3} 
                              & Random Padding + Resize (RPR)          & randomly \\
                              \hline
\multirow{1}{*}{MNIST}   & Random Padding + Resize (RPR)          & randomly \\ 
\hline
\end{tabular}
\vspace{-3mm}
\label{table:othermethods}
\end{table}

\subsection{Experimental Results}
\label{sub:ana}


\begin{table*}[!h]
\caption{The performance of variants generated by deleting mutation operators when $n$ equals 2,3,and 4. We bold the best-performing variant and underline the worst one in each group.}
\begin{tabular}{|c|llllllllllll|llllllllllll|llllllllllll|}
\hline
      Original Model           & \multicolumn{12}{c|}{MMA}                        & \multicolumn{12}{c|}{MMB}                        & \multicolumn{12}{c|}{CM}                        \\ \hline
    Acc on Testset            & \multicolumn{12}{c|}{98.56\%}                        & \multicolumn{12}{c|}{99.08\%}                        & \multicolumn{12}{c|}{79.66\%}                        \\ \hline
\multirow{2}{*}{$n=2$} & \multicolumn{6}{l}{~~~~~~\textbf{93.16\%}} & \multicolumn{6}{l|}{~~~~~~77.55\%} & \multicolumn{6}{l}{~~~~~~92.44\%} & \multicolumn{6}{l|}{~~~~~~85.36\%} & \multicolumn{6}{l}{~~~~~~71.54\%} & \multicolumn{6}{l|}{~~~~~~\textbf{72.83\%}} \\ 
                  & \multicolumn{6}{l}{~~~~~~\underline{75.09\%}} & \multicolumn{6}{l|}{~~~~~~91.65\%} & \multicolumn{6}{l}{~~~~~~\underline{77.67\%}} & \multicolumn{6}{l|}{~~~~~~\textbf{95.04\%}} & \multicolumn{6}{l}{~~~~~~\underline{68.29\%}} & \multicolumn{6}{l|}{~~~~~~69.23\%} \\ \hline
\multirow{3}{*}{$n=3$} & \multicolumn{4}{l}{~~~\textbf{98.01\%}}&\multicolumn{4}{l}{~~~86.52\%}&\multicolumn{4}{l|}{~~~97.35\%} &     \multicolumn{4}{l}{~~~\textbf{99.01\%}}&\multicolumn{4}{l}{~~~95.64\%}&\multicolumn{4}{l|}{~~~98.59\%}  &\multicolumn{4}{l}{~~~77.35\%}&\multicolumn{4}{l}{~~~75.76\%}&\multicolumn{4}{l|}{~~~\textbf{77.75\%}} \\ 
                  &   \multicolumn{4}{l}{~~~93.79\%}&\multicolumn{4}{l}{~~~\underline{75.89\%}}&\multicolumn{4}{l|}{~~~96.54\%} &     \multicolumn{4}{l}{~~~97.92\%}&\multicolumn{4}{l}{~~~\underline{79.43\%}}&\multicolumn{4}{l|}{~~~95.42\%}  &\multicolumn{4}{l}{~~~76.09\%}&\multicolumn{4}{l}{~~~\underline{71.04\%}}&\multicolumn{4}{l|}{~~~76.15\%} \\ 
                 &   \multicolumn{4}{l}{~~~97.51\%}&\multicolumn{4}{l}{~~~91.26\%}&\multicolumn{4}{l|}{~~~97.80\%} &     \multicolumn{4}{l}{~~~98.93\%}&\multicolumn{4}{l}{~~~88.16\%}&\multicolumn{4}{l|}{~~~98.60\%}  &\multicolumn{4}{l}{~~~77.08\%}&\multicolumn{4}{l}{~~~75.56\%}&\multicolumn{4}{l|}{~~~77.60\%}  \\ \hline
\multirow{4}{*}{$n=4$} &\multicolumn{3}{l}{98.41\%}&\multicolumn{3}{l}{96.18\%}&\multicolumn{3}{l}{94.30\%}&\multicolumn{3}{l|}{98.40\%}  &\multicolumn{3}{l}{99.18\%}&\multicolumn{3}{l}{98.98\%}&\multicolumn{3}{l}{98.87\%}&\multicolumn{3}{l|}{99.13\%}  &\multicolumn{3}{l}{\textbf{78.39\%}}&\multicolumn{3}{l}{77.95\%}&\multicolumn{3}{l}{78.11\%}&\multicolumn{3}{l|}{78.20\%} \\ 
                  &\multicolumn{3}{l}{98.32\%}&\multicolumn{3}{l}{89.26\%}&\multicolumn{3}{l}{\underline{86.87\%}}&\multicolumn{3}{l|}{97.92\%}  &\multicolumn{3}{l}{99.19\%}&\multicolumn{3}{l}{94.69\%}&\multicolumn{3}{l}{83.92\%}&\multicolumn{3}{l|}{98.06\%}  &\multicolumn{3}{l}{77.89\%}&\multicolumn{3}{l}{76.84\%}&\multicolumn{3}{l}{76.27\%}&\multicolumn{3}{l|}{78.21\%}        \\ 
                  &\multicolumn{3}{l}{98.05\%}&\multicolumn{3}{l}{88.74\%}&\multicolumn{3}{l}{95.67\%}&\multicolumn{3}{l|}{98.31\%}  &\multicolumn{3}{l}{99.11\%}&\multicolumn{3}{l}{\underline{82.32\%}}&\multicolumn{3}{l}{95.81\%}&\multicolumn{3}{l|}{98.77\%}  &\multicolumn{3}{l}{78.11\%}&\multicolumn{3}{l}{\underline{76.25\%}}&\multicolumn{3}{l}{77.05\%}&\multicolumn{3}{l|}{77.80\%}        \\ 
                  &\multicolumn{3}{l}{98.39\%}&\multicolumn{3}{l}{97.36\%}&\multicolumn{3}{l}{98.10\%}&\multicolumn{3}{l|}{\textbf{98.43\%}}  &\multicolumn{3}{l}{\textbf{99.21\%}}&\multicolumn{3}{l}{94.01\%}&\multicolumn{3}{l}{98.30\%}&\multicolumn{3}{l|}{99.01\%}  &\multicolumn{3}{l}{78.23\%}&\multicolumn{3}{l}{77.25\%}&\multicolumn{3}{l}{77.29\%}&\multicolumn{3}{l|}{78.07\%}     \\ \hline    
\end{tabular}
\label{table:mo}
\end{table*}

\begin{table*}[h]
\caption{The performance of MMA, MMB, and CM on newly generated test sets through image transformations when $n$ equals 2,3,4, respectively. We bold the best-performing variant and underline the worst one in each group.}
\begin{tabular}{|c|llllllllllll|llllllllllll|llllllllllll|}
\hline
       Model           & \multicolumn{12}{c|}{MMA}                        & \multicolumn{12}{c|}{MMB}                        & \multicolumn{12}{c|}{CM}                        \\ \hline
    Acc on Testset            & \multicolumn{12}{c|}{98.56\%}                        & \multicolumn{12}{c|}{99.08\%}                        & \multicolumn{12}{c|}{79.66\%}                        \\ \hline
\multirow{2}{*}{$n=2$} & \multicolumn{6}{l}{~~~~~~58.98\%} & \multicolumn{6}{l|}{~~~~~~73.01\%} & \multicolumn{6}{l}{~~~~~~\underline{59.90\%}} & \multicolumn{6}{l|}{~~~~~~80.31\%} & \multicolumn{6}{l}{~~~~~~63.83\%} & \multicolumn{6}{l|}{~~~~~~\underline{61.53\%}} \\ 
                  & \multicolumn{6}{l}{~~~~~~\textbf{70.24\%}} & \multicolumn{6}{l|}{~~~~~~\underline{54.20\%}} & \multicolumn{6}{l}{~~~~~~\textbf{88.65\%}} & \multicolumn{6}{l|}{~~~~~~74.93\%} & \multicolumn{6}{l}{~~~~~~\textbf{63.97\%}} & \multicolumn{6}{l|}{~~~~~~62.10\%} \\ \hline
\multirow{3}{*}{$n=3$} & \multicolumn{4}{l}{~~~21.65\%}&\multicolumn{4}{l}{~~~36.56\%}&\multicolumn{4}{l|}{~~~18.66\%} &     \multicolumn{4}{l}{~~~21.03\%}&\multicolumn{4}{l}{~~~41.85\%}&\multicolumn{4}{l|}{~~~29.63\%}  &\multicolumn{4}{l}{~~~42.31\%}&\multicolumn{4}{l}{~~~48.61\%}&\multicolumn{4}{l|}{~~~\underline{40.90\%}} \\ 
                  &   \multicolumn{4}{l}{~~~45.23\%}&\multicolumn{4}{l}{~~~\textbf{94.36\%}}&\multicolumn{4}{l|}{~~~43.54\%} &     \multicolumn{4}{l}{~~~56.20\%}&\multicolumn{4}{l}{~~~\textbf{96.39\%}}&\multicolumn{4}{l|}{~~~62.80\%}  &\multicolumn{4}{l}{~~~51.22\%}&\multicolumn{4}{l}{~~~\textbf{62.40\%}}&\multicolumn{4}{l|}{~~~50.88\%} \\ 
                 &   \multicolumn{4}{l}{~~~28.51\%}&\multicolumn{4}{l}{~~~39.65\%}&\multicolumn{4}{l|}{~~~\underline{11.03\%}} &     \multicolumn{4}{l}{~~~30.01\%}&\multicolumn{4}{l}{~~~42.18\%}&\multicolumn{4}{l|}{~~~\underline{10.03\%}}  &\multicolumn{4}{l}{~~~44.29\%}&\multicolumn{4}{l}{~~~55.11\%}&\multicolumn{4}{l|}{~~~45.28\%}  \\ \hline
\multirow{4}{*}{$n=4$} &\multicolumn{3}{l}{9.986\%}&\multicolumn{3}{l}{17.89\%}&\multicolumn{3}{l}{23.50\%}&\multicolumn{3}{l|}{11.09\%}  &\multicolumn{3}{l}{11.55\%}&\multicolumn{3}{l}{15.62\%}&\multicolumn{3}{l}{24.93\%}&\multicolumn{3}{l|}{13.13\%}  &\multicolumn{3}{l}{\underline{28.94\%}}&\multicolumn{3}{l}{36.62\%}&\multicolumn{3}{l}{36.80\%}&\multicolumn{3}{l|}{30.92\%} \\ 
                  &\multicolumn{3}{l}{17.98\%}&\multicolumn{3}{l}{48.33\%}&\multicolumn{3}{l}{54.86\%}&\multicolumn{3}{l|}{19.44\%}  &\multicolumn{3}{l}{27.77\%}&\multicolumn{3}{l}{51.96\%}&\multicolumn{3}{l}{56.71\%}&\multicolumn{3}{l|}{32.09\%}  &\multicolumn{3}{l}{36.80\%}&\multicolumn{3}{l}{46.37\%}&\multicolumn{3}{l}{45.72\%}&\multicolumn{3}{l|}{38.40\%}        \\ 
                  &\multicolumn{3}{l}{29.36\%}&\multicolumn{3}{l}{\textbf{55.94\%}}&\multicolumn{3}{l}{18.73\%}&\multicolumn{3}{l|}{19.66\%}  &\multicolumn{3}{l}{35.60\%}&\multicolumn{3}{l}{\textbf{76.23\%}}&\multicolumn{3}{l}{60.17\%}&\multicolumn{3}{l|}{16.39\%}  &\multicolumn{3}{l}{38.90\%}&\multicolumn{3}{l}{48.73\%}&\multicolumn{3}{l}{\textbf{49.99\%}}&\multicolumn{3}{l|}{36.96\%}        \\ 
                  &\multicolumn{3}{l}{15.09\%}&\multicolumn{3}{l}{15.18\%}&\multicolumn{3}{l}{17.01\%}&\multicolumn{3}{l|}{\underline{4.983\%}}  &\multicolumn{3}{l}{8.031\%}&\multicolumn{3}{l}{13.63\%}&\multicolumn{3}{l}{14.90\%}&\multicolumn{3}{l|}{\underline{2.997\%}}  &\multicolumn{3}{l}{35.83\%}&\multicolumn{3}{l}{48.33\%}&\multicolumn{3}{l}{44.86\%}&\multicolumn{3}{l|}{34.29\%}     \\ \hline    
\end{tabular}
\label{table:it}
\end{table*}

\begin{figure}[h]
	\centering
 \vspace{-3mm}
 \hspace{-17mm}
	\subfigure[MMA]{
		\begin{minipage}[b]{0.165\textwidth}
			\includegraphics[width=1\textwidth]{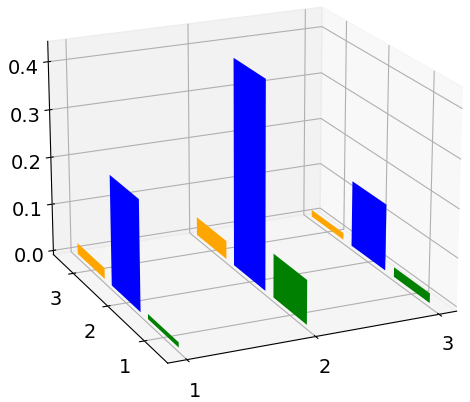}
		\end{minipage}
		\label{fig:fea1}
	}
  \hspace{-5mm}
    \subfigure[MMB]{
    	\begin{minipage}[b]{0.165\textwidth}
   		\includegraphics[width=1\textwidth]{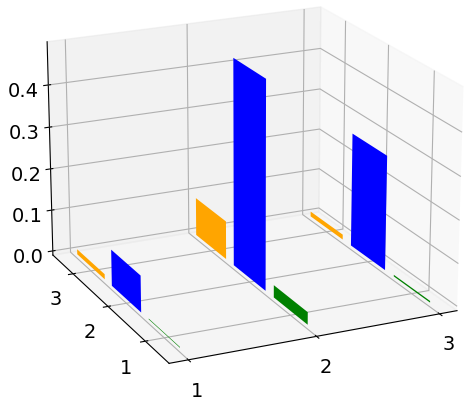}
    	\end{minipage}
	\label{fig:fea2}
    }
     \hspace{-5mm}
    \subfigure[CM]{
    	\begin{minipage}[b]{0.165\textwidth}
   		\includegraphics[width=1\textwidth]{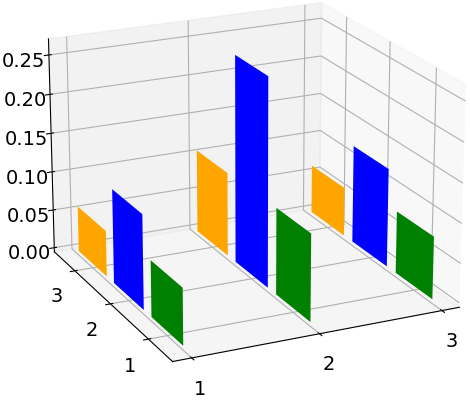}
    	\end{minipage}
	\label{fig:fea3}
    }
    \hspace{-19mm}
    \vspace{-2mm}
	\caption{The feature distribution calculated by deleting mutation operation on MMA, MMB, and CM when $n=3$. }
 \vspace{-4mm}
	\label{fig:feadisn3}
\end{figure}

\smallskip
\noindent
\textbf{Feature Distributions. }
We conduct the deleting mutation operators on MMA, MMB, and CM by respectively dividing the models into $2\times2$, $3\times3$, and $4\times4$ regions, and then deleting the corresponding region's neurons one by one.
We summarize the performance of each variant in Table~\ref{table:mo}.  Notice that the smaller the value of $n$ is, the less computation is required, but the coarser
the partition of the region is.  The second row in Table~\ref{table:mo} shows the accuracy of the original models on the original datasets. The last three rows in Table~\ref{table:mo} list the accuracy of the model variants. 
For all cases, there are performance differences among each group of variants, particularly for cases where $n=2$ and $n=3$.
Regarding MMA and MMB, the performance gap reaches approximately 20\% over regions. For the CM model, the difference is about 5\%.
Moreover, it is observed that the variants resulting from deleting the central region typically cause the most significant performance degradation, while the performance reduction of those variants deleting edge neurons is relatively minor.
We then convert the performance differences to feature distributions using~\eqref{norm} for $n=3$ and plot them in Fig.~\ref{fig:feadisn3}.
It shows that the features in both MNIST and CIFAR datasets are relatively concentrated in the central region, which aligns with our understanding of these two datasets, demonstrating the effectiveness of our approach to obtain feature distribution. 

Another observation is that in Table~\ref{table:mo}, when $n=3$, deleting neurons in the upper left region of the MMA model results in a 0.55\% decrease in accuracy, while for MMB, the decrease is only 0.07\%. We can take advantage of this result and refine the structure of CNNs.  
And for $n=4$, several variants of the MMB model with improved performance are obtained when using the deleting mutation operator, specifically by deleting the left-most column and the upper-right corner region neurons. These instances indicate that we can reduce \textbf{one-sixteenth}, even \textbf{one-ninth}, of the parameters used in all convolutional layers in this case for computational efficiency without sacrificing accuracy.

\begin{figure}[h]
	\centering
 \vspace{-3mm}
     \hspace{-17mm}
	\subfigure[MMA]{
		\begin{minipage}[b]{0.165\textwidth}
			\includegraphics[width=1\textwidth]{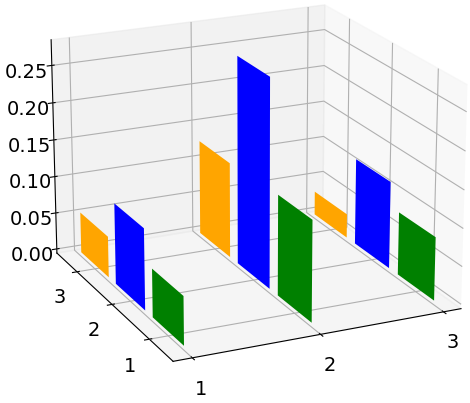}
		\end{minipage}
		\label{fig:att1}
	}    \hspace{-5mm}
    \subfigure[MMB]{
    	\begin{minipage}[b]{0.165\textwidth}
   		\includegraphics[width=1\textwidth]{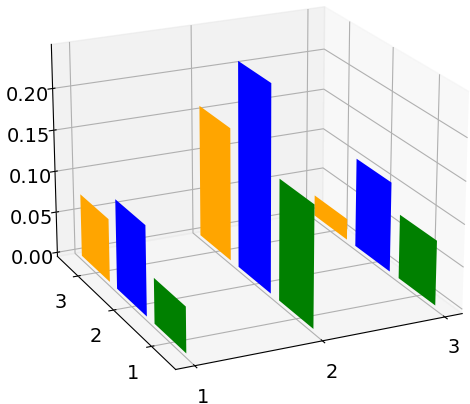}
    	\end{minipage}
	\label{fig:att2}
    }    \hspace{-5mm}
    \subfigure[CM]{
    	\begin{minipage}[b]{0.165\textwidth}
   		\includegraphics[width=1\textwidth]{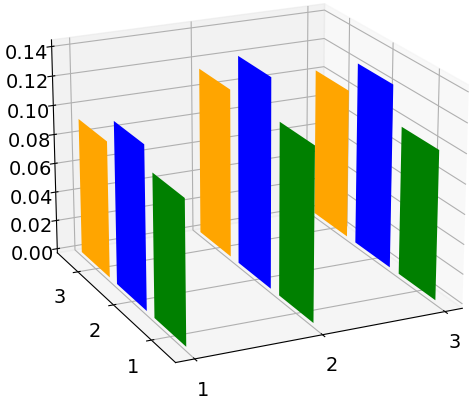}
    	\end{minipage}
	\label{fig:att3}
    }
        \hspace{-19mm}
        \vspace{-2mm}
	\caption{The attention distribution calculated by feeding transformed test sets to MMA, MMB, and CM when $n=3$. } 
 \vspace{-4mm}
	\label{fig:attdisn3}
\end{figure}

\smallskip
\noindent
\textbf{Attention Distributions.}
We generate new test sets by performing image transformation presented in Subsection~\ref{sub:score}, using values of $n$ equal to 3, 4, and 5 and setting the hyperparameter $t$ to 5.
Table~\ref{table:it} shows the accuracy of the well-trained models MMA, MMB, and CM on the generated new test sets.
We then transform these accuracy values into the models' attention allocations for each region based on~\eqref{norma} for $n=3$ and present the attention distribution in Fig.~\ref{fig:attdisn3}.
Table~\ref{table:it} indicates that all three models have a similar pattern: they exhibit higher accuracy for targets located in the central region of the image, and lower accuracy for targets located on the edges, particularly at four corners. It suggests that the model tends to pay more attention to the central region, especially for the MMA and MMB models, where the maximum difference reaches 83.33\% and 86.33\%, respectively, for $n=3$, and 50.94\% and 73.23\%, respectively, for $n=4$. In comparison, the difference in the CM model is around 20\% for $n=3$ or $4$, and only 2.44\% for $n=2$.  
This observation implies that the CM model exhibits stronger robustness with respect to the location of the object in an image. Similarly, the existence of the model's shift-invariance has also been demonstrated in this group of experiments.



\begin{table*}[htbp]
\begin{center}
\caption{The D-Scores for MMA, MMB and CM when $n$ equals 2,3, and 4, respectively. For robustness, the smaller its value, the better, while for fitness and D-Score, the larger the better. We bold the best one in each column and underline the second-best one.}
\begin{tabular}{|c|ccc|ccc|ccc|}
\hline
    & \multicolumn{3}{c|}{n=2}                                                  & \multicolumn{3}{c|}{n=3}                                                  & \multicolumn{3}{c|}{n=4}                                                  \\ \hline
    & \multicolumn{1}{c|}{$v_{\rm robust}$} & \multicolumn{1}{c|}{$v_{\rm fitness}$} & D-Score & \multicolumn{1}{c|}{$v_{\rm robust}$} & \multicolumn{1}{c|}{$v_{\rm fitness}$} & D-Score & \multicolumn{1}{c|}{$v_{\rm robust}$} & \multicolumn{1}{c|}{$v_{\rm fitness}$} & D-Score \\ \hline

MMA & \multicolumn{1}{c|}{0.2631}     & \multicolumn{1}{c|}{\underline{0.9281}}   & \underline{0.6650}  & \multicolumn{1}{c|}{0.2837}     & \multicolumn{1}{c|}{\textbf{0.9581}}   & \underline{0.6744}  & \multicolumn{1}{c|}{0.2224}     & \multicolumn{1}{c|}{\textbf{0.9728}}   & \underline{0.7504}  \\ \hline
MMB & \multicolumn{1}{c|}{\underline{0.2179}}     & \multicolumn{1}{c|}{\textbf{0.9296}}   &\textbf{0.7117}  & \multicolumn{1}{c|}{\underline{0.2758}}     & \multicolumn{1}{c|}{\underline{0.9527}}   & \textbf{0.6769}  & \multicolumn{1}{c|}{\underline{0.2202}}     & \multicolumn{1}{c|}{\underline{0.9707}}   & \textbf{0.7505}  \\ \hline
CM  & \multicolumn{1}{c|}{\textbf{0.1108}}     & \multicolumn{1}{c|}{0.7730}   & 0.6622  & \multicolumn{1}{c|}{\textbf{0.1290}}     & \multicolumn{1}{c|}{0.7813}   & 0.6523  & \multicolumn{1}{c|}{\textbf{0.1083}}     & \multicolumn{1}{c|}{0.7933}   & 0.6849  \\ \hline
\end{tabular}
\end{center}
\label{table:score3}
\end{table*}

\begin{table*}[h]
\begin{center}
\caption{The D-Score for CM with different data augmentation methods. We bold the best score and underline the second-best one.}
\vspace{-2mm}
\begin{tabular}{|c|lll|lll|lll|}
\hline
Methods                                                                                            & \multicolumn{3}{c|}{$a_i$} & \multicolumn{3}{c|}{$\widetilde{f}_i$} 
                                                                                                                                                       & \multicolumn{1}{l|}{$v_{\rm
                                                        robust}$($\downarrow$ better)}              & \multicolumn{1}{l|}{$v_{\rm fitness}$ ($\uparrow$ better)}                & D-Score ($\uparrow$ better)                \\ \hline
                                                                                                                    
  \multirow{3}{*}{\begin{tabular}[c]{@{}c@{}}w/o Augmentation\\ Loss:0.5694\\ \underline{Acc:79.66\%}\end{tabular}}                            & 42.31\%          & 48.61\%          & 40.90\%         & 7.095\%           &11.98\%          & 5.866\%          & \multicolumn{1}{l|}{\multirow{3}{*}{0.1290}} & \multicolumn{1}{l|}{\multirow{3}{*}{0.7813}} & \multirow{3}{*}{0.6523} \\
                                                                                                                    & 51.22\%          & 62.40\%          & 50.88\%         & 10.96\%           & 26.47\%          & 10.78\%          & \multicolumn{1}{l|}{}                        & \multicolumn{1}{l|}{}                        &                         \\
                                                                                                                    & 44.29\%          & 55.11\%          & 45.28\%         & 7.924\%           & 12.59\%          & 6.327\%          & \multicolumn{1}{l|}{}                        & \multicolumn{1}{l|}{}                        &                         \\ \hline                                                                                                                  
                                                                                                                    
\multirow{3}{*}{\begin{tabular}[c]{@{}c@{}}RHF\\ Loss:0.5694\\ \textbf{Acc:81.22\%}\end{tabular}}                            & 45.32\%          & 52.10\%          & 45.11\%         & 3.543\%           & 10.41\%          & 6.011\%          & \multicolumn{1}{l|}{\multirow{3}{*}{0.1221}} & \multicolumn{1}{l|}{\multirow{3}{*}{\textbf{0.7934}}} & \multirow{3}{*}{\underline{0.6713}} \\
                                                                                                                    & 56.63\%          & 66.98\%          & 57.91\%         & 6.634\%           & 25.37\%          & 8.257\%          & \multicolumn{1}{l|}{}                        & \multicolumn{1}{l|}{}                        &                         \\
                                                                                                                    & 49.10\%          & 58.23\%          & 50.17\%         & 9.870\%           & 20.40\%          & 9.491\%          & \multicolumn{1}{l|}{}                        & \multicolumn{1}{l|}{}                        &                         \\ \hline
\multirow{3}{*}{\begin{tabular}[c]{@{}c@{}}RVF\\ Loss:0.7573\\ Acc:73.70\%\end{tabular}}                            & 40.99\%          & 47.32\%          & 36.68\%         & 2.678\%           & 12.51\%          & 4.407\%          & \multicolumn{1}{l|}{\multirow{3}{*}{0.1265}} & \multicolumn{1}{l|}{\multirow{3}{*}{0.7144}} & \multirow{3}{*}{0.5878} \\
                                                                                                                    & 50.43\%          & 58.73\%          & 45.98\%         & 10.64\%           & 30.10\%          & 9.152\%          & \multicolumn{1}{l|}{}                        & \multicolumn{1}{l|}{}                        &                         \\
                                                                                                                    & 44.33\%          & 49.17\%          & 40.50\%         & 8.576\%           & 18.20\%          & 3.723\%          & \multicolumn{1}{l|}{}                        & \multicolumn{1}{l|}{}                        &                         \\ \hline
\multirow{3}{*}{\begin{tabular}[c]{@{}c@{}}RR\\ Loss:0.9375\\ Acc:67.10\%\end{tabular}}                             & 32.10\%          & 38.17\%          & 27.98\%         & 5.390\%           & 7.338\%          & 7.009\%          & \multicolumn{1}{l|}{\multirow{3}{*}{0.1302}} & \multicolumn{1}{l|}{\multirow{3}{*}{0.6576}} & \multirow{3}{*}{0.5274} \\
                                                                                                                    & 41.23\%          & 45.96\%          & 34.47\%         & 10.96\%           & 22.60\%          & 15.67\%          & \multicolumn{1}{l|}{}                        & \multicolumn{1}{l|}{}                        &                         \\
                                                                                                                    & 35.60\%          & 37.88\%          & 26.35\%         & 12.65\%           & 11.36\%          & 7.010\%          & \multicolumn{1}{l|}{}                        & \multicolumn{1}{l|}{}                        &                         \\ \hline
\multirow{3}{*}{\begin{tabular}[c]{@{}c@{}}RHV\\ Loss:0.7038\\ Acc:75.66\%\end{tabular}}                            & 38.01\%          & 45.11\%          & 39.81\%         & 4.801\%           & 12.03\%          & 8.647\%          & \multicolumn{1}{l|}{\multirow{3}{*}{\underline{0.1219}}} & \multicolumn{1}{l|}{\multirow{3}{*}{0.7458}} & \multirow{3}{*}{0.6240} \\
                                                                                                                    & 47.30\%          & 57.63\%          & 47.83\%         & 7.408\%           & 20.65\%          & 10.53\%          & \multicolumn{1}{l|}{}                        & \multicolumn{1}{l|}{}                        &                         \\
                                                                                                                    & 43.26\%          & 49.71\%          & 39.99\%         & 11.69\%           & 15.02\%          & 9.215\%          & \multicolumn{1}{l|}{}                        & \multicolumn{1}{l|}{}                        &                         \\ \hline
\multirow{3}{*}{\begin{tabular}[c]{@{}c@{}}Ours (p=0.26)\\ Loss:0.6274\\ Acc:79.44\%\end{tabular}}             &      49.98\%                     &     61.12\%             &       50.01\%           &        7.117\%         &    11.84\%               &      7.056\%                             & \multicolumn{1}{l|}{\multirow{3}{*}{\textbf{0.0865}}}       & \multicolumn{1}{l|}{\multirow{3}{*}{\underline{0.7839}}}       & \multirow{3}{*}{\textbf{0.6974}}       \\
    &   66.32\%       &      76.42\%            &     64.33\%             & 10.17\%                &  21.95\%                 &        9.448\%                           & \multicolumn{1}{l|}{}                        & \multicolumn{1}{l|}{}                        &                      \\
    
                                                                                                                   &57.04\%                  &     66.08\%             &      56.97\%           &       8.177\%            &       13.51\%           &      10.72\%            & \multicolumn{1}{l|}{}                        & \multicolumn{1}{l|}{}                        &                         \\ \hline
\end{tabular}
\vspace{-3mm}
\label{table:cifarAug}
\end{center}
\end{table*}

\begin{figure*}[ht]
  \centering
   \includegraphics[width=0.95\textwidth]{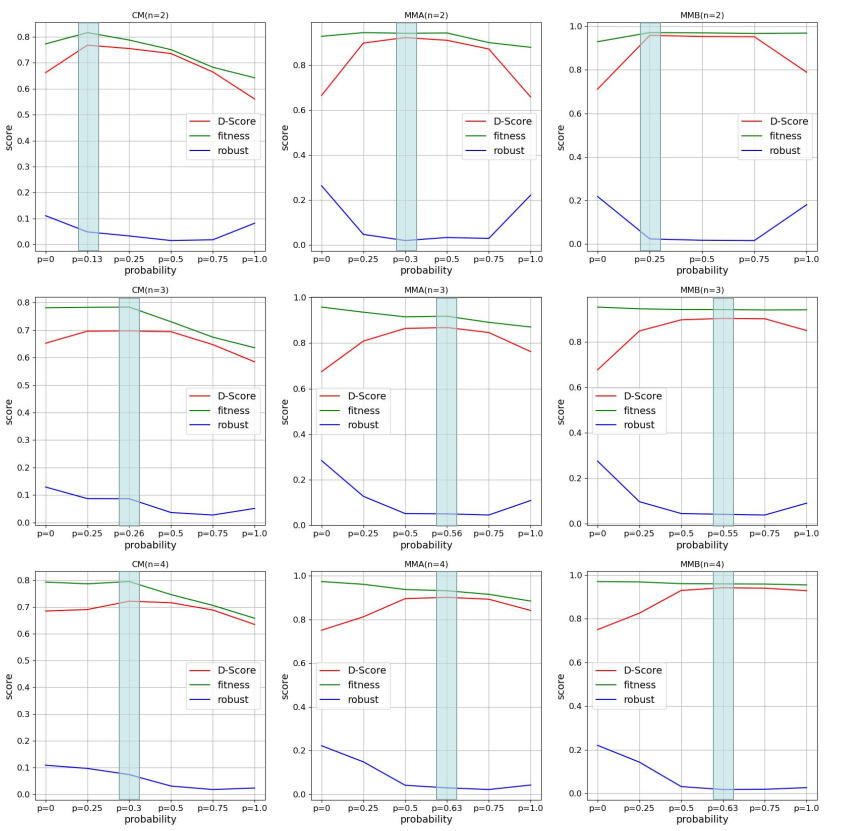}
    \caption{The D-Scores of CM, MMA, and MMB after applying our data augmentation method with different probabilities. It is evident that the highest D-Score is obtained by using the execution probability calculated based on $p=\frac{v_{\rm robust}}{g(n)}$, achieving a good balance between robustness and fitness. The shaded area indicates the range where the transitions of D-Scores from increasing to decreasing, which is consistent in all three models and two datasets, demonstrating the effectiveness of our score-guiding method and the rationality of D-Score in assessing CNN fitness and robustness.
    }
	\label{scoredifferentP}
\end{figure*}

\smallskip
\noindent
\textbf{Scoring.}
Combining the feature distribution and the attention distribution, we calculate D-Scores for the three models based on~\eqref{score}, as shown in Table~\ref{table:score3}.
Take $n=3$ as an example, for robustness, the CM model performs the best with a score of 0.1290 (the lower the better), while MMA and MMB have scores of 0.2837 and 0.2758, respectively. This aligns with our understanding of the two datasets and the results in our previous experiments: the MNIST dataset primarily consists of digits in the center of the image, which leads to models trained on this dataset being insensitive to features located on the edges of the image and unable to recognize corner cases. Although the CIFAR dataset suffers from the same problem, it is less severe due to the more complex nature of the images, where the features are distributed in multiple locations besides the center of the image.
Therefore, from the robustness perspective, the CM model demonstrates a more consistent ability to recognize features in all regions compared to the other two models. However, in terms of fitness, after calculating the difference in feature and attention distribution, MMB and MMA still outperform CM due to CM's lower accuracy on the original test set.
Then we obtain the final D-Score.
Based on our evaluation, we believe that the MMB model performs the best overall, followed by MMA, and the choice of $n$ has no effect on the relative ranking of the D-Scores for the three models.

\smallskip
\noindent
\textbf{Score-guided Augmentation.}
Both MNIST and CIFAR-10 contain 10 classes, which means $c=10$. According to~\eqref{eq:fn}, we then perform the proposed score-guided data augmentation for $n = 2, 3, 4$, with the corresponding values of $g(n)$ and the related execution probability $p$, which are summarized in Table~\ref{table:para}.
\begin{table}[]
\begin{center}
\caption{The values of $g(n)$ and the probability $p$ when $n$ equals 2, 3, 4.}
\begin{tabular}{|c|c|c|c|}
\hline
     & n=2  & n=3  & n=4  \\ \hline
$g(n)$ & 0.88 & 0.50 & 0.35 \\ \hline
$p_{\text{CM}}$   & 0.13 & 0.26 & 0.30 \\ \hline
$p_{\text{MMA}}$  & 0.30 & 0.56 & 0.63 \\ \hline
$p_{\text{MMB}}$  & 0.25 & 0.55 & 0.63 \\ \hline
\end{tabular}
\end{center}
\label{table:para}
\end{table}

Take $n=3$ as an example, we apply our data augmentation method to CIFAR-10 and compare it with other approaches, including RHF, RVF, RR, RHV and the case without augmentation.
The results are summarized in Table~\ref{table:cifarAug}.
Our method achieves the best results in D-Score and $v_{\rm robust}$, and the second-best result in $v_{\rm fitness}$. Compared to the method without any data augmentation, our method significantly improves the robustness of the model while maintaining a similar level of accuracy (only 0.22\% difference in the first column). The robustness can be clearly observed in the ``$a_i$'' column, where the model accuracy for objects appearing in different regions significantly increases (e.g., the accuracy for objects in the lower right corner increased from 45.28\% to 56.97\%.
We notice that the RHF method achieves the best accuracy score (81.22\%) and the highest fitness score (0.7934) in the test set, which shows the consistency between accuracy and our proposed fitness. From the ``$a_i$'' column, we can see that using our scheme the model's accuracy scores for various regions are significantly improved. So are the feature distributions of the dataset (features are more equally distributed), as shown in the ``$\widetilde{f}_i$'' column. For example, the top left corner of RHF only has 3.543\% of the features, while the middle region has 25.37\%, which is unbalanced. In contrast, through our method, the feature content in the top left corner increases to 7.117\%, while the middle area decreases to 21.95\%.

Another experiment is to use our augmentation method with different $p$ on MNIST and CIFAR-10 to show the impact of the execution probability on the scores. The results are shown in Fig.~\ref{scoredifferentP}.
Clearly, using our proposed $p$ in~\eqref{eq:p}, based on the robustness score, yields the highest D-Score in all cases. Moreover, this calculated probability falls exactly in the interval where the D-Score transitions from increasing to decreasing in all cases, as indicated by the masked areas. 

\section{Conclusions}
\label{con}
This paper studies how to effectively evaluate robustness of CNNs and their fitness to datasets, rather than just rely on their scores on an unevaluated test set.
We propose a white-box method for this purpose, which analyzes the feature distribution of the dataset and the attention distribution of the model, using mutation operators and well-designed image transformations, respectively.  With these distributions, we introduce D-Score to reflect the model's robustness and fitness. 
To demonstrate that our score can effectively represent the robustness of CNNs, we propose a score-guided data augmentation method to address the issue of CNN's lack of translation invariance. We validate our approach on two widely used datasets and three widely adopted models.

It is interesting to observe from Table~\ref{table:mo} that removing certain regions of neurons does not adversely affect the classification results.  In fact, it sometimes even improves performance. So our analysis can potentially provide a guideline for CNN structure refinement and parameter reduction, which is different from the traditional dropout approach~\cite{srivastava2014dropout}, which requires to set the ratio of dropout before training.

\bibliographystyle{IEEEtran}
\bibliography{reference}

\begin{thebibliography}{10}
\providecommand{\url}[1]{#1}
\csname url@samestyle\endcsname
\providecommand{\newblock}{\relax}
\providecommand{\bibinfo}[2]{#2}
\providecommand{\BIBentrySTDinterwordspacing}{\spaceskip=0pt\relax}
\providecommand{\BIBentryALTinterwordstretchfactor}{4}
\providecommand{\BIBentryALTinterwordspacing}{\spaceskip=\fontdimen2\font plus
\BIBentryALTinterwordstretchfactor\fontdimen3\font minus
  \fontdimen4\font\relax}
\providecommand{\BIBforeignlanguage}[2]{{%
\expandafter\ifx\csname l@#1\endcsname\relax
\typeout{** WARNING: IEEEtran.bst: No hyphenation pattern has been}%
\typeout{** loaded for the language `#1'. Using the pattern for}%
\typeout{** the default language instead.}%
\else
\language=\csname l@#1\endcsname
\fi
#2}}
\providecommand{\BIBdecl}{\relax}
\BIBdecl

\bibitem{litjens2017survey}
G.~Litjens, T.~Kooi, B.~E. Bejnordi, A.~A.~A. Setio, F.~Ciompi, M.~Ghafoorian,
  J.~A. Van Der~Laak, B.~Van~Ginneken, and C.~I. S{\'a}nchez, ``A survey on
  deep learning in medical image analysis,'' \emph{Medical image analysis},
  vol.~42, pp. 60--88, 2017.

\bibitem{ozturk2020automated}
T.~Ozturk, M.~Talo, E.~A. Yildirim, U.~B. Baloglu, O.~Yildirim, and U.~R.
  Acharya, ``Automated detection of covid-19 cases using deep neural networks
  with x-ray images,'' \emph{Computers in biology and medicine}, vol. 121, p.
  103792, 2020.

\bibitem{grigorescu2020survey}
S.~Grigorescu, B.~Trasnea, T.~Cocias, and G.~Macesanu, ``A survey of deep
  learning techniques for autonomous driving,'' \emph{Journal of Field
  Robotics}, vol.~37, no.~3, pp. 362--386, 2020.

\bibitem{gupta2021deep}
A.~Gupta, A.~Anpalagan, L.~Guan, and A.~S. Khwaja, ``Deep learning for object
  detection and scene perception in self-driving cars: Survey, challenges, and
  open issues,'' \emph{Array}, vol.~10, p. 100057, 2021.

\bibitem{denos2017deep}
K.~Denos, M.~Ravaut, A.~Fagette, and H.-S. Lim, ``Deep learning applied to
  underwater mine warfare,'' in \emph{OCEANS 2017-Aberdeen}.\hskip 1em plus
  0.5em minus 0.4em\relax IEEE, 2017, pp. 1--7.

\bibitem{kafedziski2018detection}
V.~Kafedziski, S.~Pecov, and D.~Tanevski, ``Detection and classification of
  land mines from ground penetrating radar data using faster r-cnn,'' in
  \emph{2018 26th telecommunications forum (TELFOR)}.\hskip 1em plus 0.5em
  minus 0.4em\relax IEEE, 2018, pp. 1--4.

\bibitem{feng2020deepgini}
Y.~Feng, Q.~Shi, X.~Gao, J.~Wan, C.~Fang, and Z.~Chen, ``Deepgini: prioritizing
  massive tests to enhance the robustness of deep neural networks,'' in
  \emph{Proceedings of the 29th ACM SIGSOFT International Symposium on Software
  Testing and Analysis}, 2020, pp. 177--188.

\bibitem{zhang2020machine}
J.~M. Zhang, M.~Harman, L.~Ma, and Y.~Liu, ``Machine learning testing: Survey,
  landscapes and horizons,'' \emph{IEEE Transactions on Software Engineering},
  2020.

\bibitem{ma2018deepmutation}
L.~Ma, F.~Zhang, J.~Sun, M.~Xue, B.~Li, F.~Juefei-Xu, C.~Xie, L.~Li, Y.~Liu,
  J.~Zhao \emph{et~al.}, ``Deepmutation: Mutation testing of deep learning
  systems,'' in \emph{2018 IEEE 29th International Symposium on Software
  Reliability Engineering (ISSRE)}.\hskip 1em plus 0.5em minus 0.4em\relax
  IEEE, 2018, pp. 100--111.

\bibitem{sun2018testing}
Y.~Sun, X.~Huang, D.~Kroening, J.~Sharp, M.~Hill, and R.~Ashmore, ``Testing
  deep neural networks,'' \emph{arXiv preprint arXiv:1803.04792}, 2018.

\bibitem{pei2017deepxplore}
K.~Pei, Y.~Cao, J.~Yang, and S.~Jana, ``Deepxplore: Automated whitebox testing
  of deep learning systems,'' in \emph{proceedings of the 26th Symposium on
  Operating Systems Principles}, 2017, pp. 1--18.

\bibitem{hu2019deepmutation++}
Q.~Hu, L.~Ma, X.~Xie, B.~Yu, Y.~Liu, and J.~Zhao, ``Deepmutation++: A mutation
  testing framework for deep learning systems,'' in \emph{2019 34th IEEE/ACM
  International Conference on Automated Software Engineering (ASE)}.\hskip 1em
  plus 0.5em minus 0.4em\relax IEEE, 2019, pp. 1158--1161.

\bibitem{humbatova2021deepcrime}
N.~Humbatova, G.~Jahangirova, and P.~Tonella, ``Deepcrime: mutation testing of
  deep learning systems based on real faults,'' in \emph{Proceedings of the
  30th ACM SIGSOFT International Symposium on Software Testing and Analysis},
  2021, pp. 67--78.

\bibitem{shen2018munn}
W.~Shen, J.~Wan, and Z.~Chen, ``Munn: Mutation analysis of neural networks,''
  in \emph{2018 IEEE International Conference on Software Quality, Reliability
  and Security Companion (QRS-C)}.\hskip 1em plus 0.5em minus 0.4em\relax IEEE,
  2018, pp. 108--115.

\bibitem{panichella2021we}
A.~Panichella and C.~C. Liem, ``What are we really testing in mutation testing
  for machine learning? a critical reflection,'' in \emph{2021 IEEE/ACM 43rd
  International Conference on Software Engineering: New Ideas and Emerging
  Results (ICSE-NIER)}.\hskip 1em plus 0.5em minus 0.4em\relax IEEE, 2021, pp.
  66--70.

\bibitem{yu2019test4deep}
J.~Yu, Y.~Fu, Y.~Zheng, Z.~Wang, and X.~Ye, ``Test4deep: an effective white-box
  testing for deep neural networks,'' in \emph{2019 IEEE international
  conference on computational science and engineering (CSE) and IEEE
  international conference on embedded and ubiquitous computing (EUC)}.\hskip
  1em plus 0.5em minus 0.4em\relax IEEE, 2019, pp. 16--23.

\bibitem{harel2020neuron}
F.~Harel-Canada, L.~Wang, M.~A. Gulzar, Q.~Gu, and M.~Kim, ``Is neuron coverage
  a meaningful measure for testing deep neural networks?'' in \emph{Proceedings
  of the 28th ACM Joint Meeting on European Software Engineering Conference and
  Symposium on the Foundations of Software Engineering}, 2020, pp. 851--862.

\bibitem{small}
A.~Azulay and Y.~Weiss, ``Why do deep convolutional networks generalize so
  poorly to small image transformations?'' \emph{arXiv preprint
  arXiv:1805.12177}, 2018.

\bibitem{zhang2019making}
R.~Zhang, ``Making convolutional networks shift-invariant again,'' in
  \emph{International conference on machine learning}.\hskip 1em plus 0.5em
  minus 0.4em\relax PMLR, 2019, pp. 7324--7334.

\bibitem{shorten2019survey}
C.~Shorten and T.~M. Khoshgoftaar, ``A survey on image data augmentation for
  deep learning,'' \emph{Journal of big data}, vol.~6, no.~1, pp. 1--48, 2019.

\bibitem{offutt2001mutation}
A.~J. Offutt and R.~H. Untch, ``Mutation 2000: Uniting the orthogonal,''
  \emph{Mutation testing for the new century}, pp. 34--44, 2001.

\bibitem{ma2006mujava}
Y.-S. Ma, J.~Offutt, and Y.-R. Kwon, ``Mujava: a mutation system for java,'' in
  \emph{Proceedings of the 28th international conference on Software
  engineering}, 2006, pp. 827--830.

\bibitem{papadakis2019mutation}
M.~Papadakis, M.~Kintis, J.~Zhang, Y.~Jia, Y.~Le~Traon, and M.~Harman,
  ``Mutation testing advances: an analysis and survey,'' in \emph{Advances in
  Computers}.\hskip 1em plus 0.5em minus 0.4em\relax Elsevier, 2019, vol. 112,
  pp. 275--378.

\bibitem{agrawal1989design}
H.~Agrawal, R.~A. DeMillo, R.~Hathaway, W.~Hsu, W.~Hsu, E.~W. Krauser, R.~J.
  Martin, A.~P. Mathur, and E.~Spafford, ``Design of mutant operators for the c
  programming language,'' Citeseer, Tech. Rep., 1989.

\bibitem{jia2010analysis}
Y.~Jia and M.~Harman, ``An analysis and survey of the development of mutation
  testing,'' \emph{IEEE transactions on software engineering}, vol.~37, no.~5,
  pp. 649--678, 2010.

\bibitem{wang2019adversarial}
J.~Wang, G.~Dong, J.~Sun, X.~Wang, and P.~Zhang, ``Adversarial sample detection
  for deep neural network through model mutation testing,'' in \emph{2019
  IEEE/ACM 41st International Conference on Software Engineering (ICSE)}.\hskip
  1em plus 0.5em minus 0.4em\relax IEEE, 2019, pp. 1245--1256.

\bibitem{jahangirova2020empirical}
G.~Jahangirova and P.~Tonella, ``An empirical evaluation of mutation operators
  for deep learning systems,'' in \emph{2020 IEEE 13th International Conference
  on Software Testing, Validation and Verification (ICST)}.\hskip 1em plus
  0.5em minus 0.4em\relax IEEE, 2020, pp. 74--84.

\bibitem{gannamaneni2022good}
S.~S. Gannamaneni, M.~Akila, C.~Heinzemann, and M.~Woehrle, ``The good and the
  bad: using neuron coverage as a dnn validation technique,'' in \emph{Deep
  Neural Networks and Data for Automated Driving: Robustness, Uncertainty
  Quantification, and Insights Towards Safety}.\hskip 1em plus 0.5em minus
  0.4em\relax Springer International Publishing Cham, 2022, pp. 383--403.

\bibitem{albawi2017understanding}
S.~Albawi, T.~A. Mohammed, and S.~Al-Zawi, ``Understanding of a convolutional
  neural network,'' in \emph{2017 international conference on engineering and
  technology (ICET)}.\hskip 1em plus 0.5em minus 0.4em\relax Ieee, 2017, pp.
  1--6.

\bibitem{o2015introduction}
K.~O'Shea and R.~Nash, ``An introduction to convolutional neural networks,''
  \emph{arXiv preprint arXiv:1511.08458}, 2015.

\bibitem{wu2017introduction}
J.~Wu, ``Introduction to convolutional neural networks,'' \emph{National Key
  Lab for Novel Software Technology. Nanjing University. China}, vol.~5,
  no.~23, p. 495, 2017.

\bibitem{fukushima1980neocognitron}
K.~Fukushima, ``Neocognitron: A self-organizing neural network model for a
  mechanism of pattern recognition unaffected by shift in position,''
  \emph{Biological cybernetics}, vol.~36, no.~4, pp. 193--202, 1980.

\bibitem{fukushima1982neocognitron}
K.~Fukushima and S.~Miyake, ``Neocognitron: A new algorithm for pattern
  recognition tolerant of deformations and shifts in position,'' \emph{Pattern
  recognition}, vol.~15, no.~6, pp. 455--469, 1982.

\bibitem{sp}
M.~Jaderberg, K.~Simonyan, A.~Zisserman \emph{et~al.}, ``Spatial transformer
  networks,'' \emph{Advances in neural information processing systems},
  vol.~28, 2015.

\bibitem{deng2012mnist}
L.~Deng, ``The mnist database of handwritten digit images for machine learning
  research [best of the web],'' \emph{IEEE signal processing magazine},
  vol.~29, no.~6, pp. 141--142, 2012.

\bibitem{krizhevsky2017imagenet}
A.~Krizhevsky, I.~Sutskever, and G.~E. Hinton, ``Imagenet classification with
  deep convolutional neural networks,'' \emph{Communications of the ACM},
  vol.~60, no.~6, pp. 84--90, 2017.

\bibitem{lecun1998gradient}
Y.~LeCun, L.~Bottou, Y.~Bengio, and P.~Haffner, ``Gradient-based learning
  applied to document recognition,'' \emph{Proceedings of the IEEE}, vol.~86,
  no.~11, pp. 2278--2324, 1998.

\bibitem{xiao2018generating}
C.~Xiao, B.~Li, J.-Y. Zhu, W.~He, M.~Liu, and D.~Song, ``Generating adversarial
  examples with adversarial networks,'' \emph{arXiv preprint arXiv:1801.02610},
  2018.

\bibitem{carlini2017towards}
N.~Carlini and D.~Wagner, ``Towards evaluating the robustness of neural
  networks,'' in \emph{2017 ieee symposium on security and privacy (sp)}.\hskip
  1em plus 0.5em minus 0.4em\relax Ieee, 2017, pp. 39--57.

\bibitem{srivastava2014dropout}
N.~Srivastava, G.~Hinton, A.~Krizhevsky, I.~Sutskever, and R.~Salakhutdinov,
  ``Dropout: a simple way to prevent neural networks from overfitting,''
  \emph{The journal of machine learning research}, vol.~15, no.~1, pp.
  1929--1958, 2014.

\end{thebibliography}

\end{document}